\documentclass[sn-mathphys-num]{sn-jnl}% Math and Physical Sciences Numbered Reference Style
\usepackage{graphicx}%
\usepackage{multirow}%
\usepackage{amsmath,amssymb,amsfonts}%
\usepackage{amsthm}%
\usepackage{mathrsfs}%
\usepackage[title]{appendix}%
\usepackage{xcolor}%
\usepackage{textcomp}%
\usepackage{manyfoot}%
\usepackage{booktabs}%
\usepackage{algorithm}%
\usepackage{algorithmicx}%
\usepackage{algpseudocode}%
\usepackage{listings}%
\usepackage{float}%
\theoremstyle{thmstyleone}%
\newtheorem{theorem}{Theorem}%  meant for continuous numbers
\newtheorem{proposition}[theorem]{Proposition}% 

\theoremstyle{thmstyletwo}%
\newtheorem{example}{Example}%
\newtheorem{remark}{Remark}%

\theoremstyle{thmstylethree}%
\newtheorem{definition}{Definition}%

\begin{document}

\title[Article Title]{Quantifying Information Loss under Coarse-Grained Partitions: A Discrete Framework for Explainable Artificial Intelligence}


\author*[1]{\fnm{Takashi} \sur{Izumo}}\email{izumo.takashi@nihon-u.ac.jp}

\affil*[1]{\orgdiv{College of Law}, \orgname{Nihon University}, \orgaddress{\street{Kandamisakimachi}, \city{Chiyodaku}, \postcode{1018351}, \state{Tokyo}, \country{Japan}, ORCID: 0000-0003-0008-4729}}


\abstract{
As artificial intelligence (AI) systems are increasingly used in ethically sensitive domains such as education, healthcare, and transportation, balancing accuracy and interpretability has become a central concern. Coarse ethics (CE) motivates coarse-grained evaluations under cognitive, institutional, and contextual constraints, but it still lacks a simple mathematical formalization of admissible coarse-graining and its informational consequences. This paper introduces coarse-grained partitions (CGPs) as a discrete framework for modeling coarse evaluation on a finite totally ordered score scale. A CGP represents coarse evaluation as a partition into grains with an index assignment, and induces a coarse-grained distribution by pushforward. To compare admissible coarse-grainings, we introduce categorical unification (CU), which constructs a canonical fine-scale reconstruction from the coarse representation under minimal assumptions. On this basis, we define a KL-based measure of information loss, $D_{\mathrm{KL\text{-}CU}}$, as the divergence between the original fine-grained distribution and its CU-based reconstruction. We prove that $D_{\mathrm{KL\text{-}CU}}=0$ if and only if the original distribution is already uniform within each grain. This shows that zero loss, in the sense of the proposed measure, is a highly exceptional limiting case rather than a realistic benchmark for ordinary evaluative practice. We also show that the framework leads naturally to an optimization problem for comparing alternative admissible CGPs. Applications to educational grading and explainable AI (XAI) illustrate how the framework clarifies trade-offs among informational fidelity, interpretability, and coarsening cost.
}




\maketitle

\section{Introduction}\label{sec1}
\subsection{Background}\label{sec1.1}
As artificial intelligence (AI) becomes increasingly embedded in decision-making processes, the need for effective user interaction with AI-driven systems has become a critical research concern. Concepts such as fairness, transparency, and explainability, including explainable AI (XAI), have been central to this discourse \citep{gunning2019xai}. However, a well-established trade-off exists between predictive accuracy and interpretability in AI models, where higher precision often results in reduced explainability \citep{BARREDOARRIETA202082}.

To address this challenge, researchers have explored various approximation techniques, which can be broadly categorized into engineering-driven and ethics-driven approaches. From an engineering perspective, a common strategy involves approximating complex black-box models with more interpretable alternatives. Techniques such as Local Interpretable Model-Agnostic Explanations (LIME, \cite{ribeiro2016whyitrustyou}) and SHapley Additive exPlanations (SHAP, \cite{lundberg2017unifiedapproachinterpretingmodel}) have been widely used to provide post-hoc interpretability for AI systems. 

Recent research on AI-assisted decision-making has increasingly emphasized that explanation and evaluation are not purely technical matters, but also ethical ones. In particular, a growing body of work has examined how AI outputs should be presented to human users in ways that are cognitively manageable, socially appropriate, and normatively defensible. Several prior studies are relevant in this respect. For example, He et al.~\cite{he2024analogies} explored how analogies can assist laypeople in AI-assisted decision-making, highlighting the cognitive conditions under which explanations become understandable to non-expert users. Peters and Carman~\cite{peters2024cultural} examined the influence of cultural bias in XAI research, showing that users from different cultural backgrounds may interpret AI explanations differently. These studies underscore that the form and granularity of AI explanations are not neutral, but depend on the capacities, expectations, and interpretive contexts of their audiences.

From this broader ethical perspective, coarse ethics (CE, \cite{izumoweng2022}, \citep{izumo2025}) offers a particularly suggestive framework. CE proposes that coarse-grained assessments are not only practically unavoidable but also ethically justifiable in many decision-making contexts. For example, academic grading systems classify students into letter grades or grade point averages (GPAs), thereby abstracting away precise score differences, and expressions such as ``Olivia and Noah are Harvard graduates'' similarly suppress finer distinctions in individual performance. In AI-driven decision-making, CE argues that the legitimacy of transforming fine-grained evaluations into coarse ones depends on the cognitive capacity and contextual needs of the audience \citep{izumo2025}.

The present paper focuses on this CE perspective. Despite its conceptual appeal, CE faces a central challenge: it lacks a mathematically rigorous structure, and this weakens the usefulness of the very notion of coarse-grained normative evaluation. Izumo and Weng \cite{izumoweng2022} identify two requirements for coarsening a fine-grained evaluation: first, that the coarse-grained evaluation should sufficiently cover the objects evaluated in the original one (for example, if Olivia is evaluated in the original assessment, she should not disappear from the coarse-grained version); and second, that the coarse-grained evaluation should not reverse the value ordering embodied in the fine-grained one (for example, if Olivia is rated higher than Noah in the original evaluation, the coarsened evaluation should not end up ranking Noah above Olivia). However, as the authors themselves acknowledge, these conditions do not suffice to determine a unique coarse-grained evaluation. As a result, multiple distinct coarse evaluations satisfying those conditions may arise from a single fine-grained one. This problem stems from the fact that the previous literature does not specify the mathematical process by which coarsening is carried out.

\subsection{Methodology}\label{sec1.2}

This paper proposes coarse-grained partitions (CGPs) as a novel set-theoretic framework for addressing these issues. Set-theoretic approaches have long been used in the study of human decision-making, most notably in fuzzy set theory \citep{zadeh1965fuzzy}, rough set theory \citep{pawlak1982rough}, and soft set theory \citep{molodtsov1999soft}. These approaches were primarily developed to represent uncertainty, vagueness, and incomplete information in the world as mathematically tractable structures. CGPs differ from these approaches in their point of departure. Rather than improving AI's ability to model external reality, CGPs are designed to analyze how fine-grained evaluations can be transformed into coarser forms that remain usable for human agents. In this sense, the problem is no longer how AI perceives the world, but how human beings can understand, interpret, and act upon AI-generated evaluations. This perspective becomes especially important in domains where AI systems now outperform human experts in narrowly defined tasks, including standardized testing and other forms of specialized assessment \citep{bicknell2024chatgpt, katz2023gpt4}. As AI systems become increasingly capable, the challenge of translating their internal evaluations into forms intelligible to human users becomes correspondingly more urgent.

From this perspective, CGPs provide a mathematical framework for studying coarse-grained evaluation itself. We formalize the transformation from fine-grained to coarse-grained evaluation by introducing partitions of a finite discrete totally ordered score scale. We then compare the probability distributions before and after coarsening, define a canonical reconstruction of the fine-grained distribution through categorical unification, and measure information loss by KL divergence \citep{kullbackleibler1951} with categorical unification (CU), i.e., $D_\mathrm{{KL\text{-}CU}}$. This allows us to analyze, in precise terms, when coarse-graining preserves information, when it necessarily destroys it, and how alternative coarse-graining schemes may be compared. More precisely, the role of CGPs in this paper is twofold. First, as interval partitions of a finite totally ordered scale, they provide a mathematically explicit class of coarse-grainings that satisfies the structural requirements emphasized in CE, namely coverage and order preservation. Second, since these requirements alone do not determine a unique coarse-graining, we introduce $D_\mathrm{{KL\text{-}CU}}$ not as a replacement for them, but as an additional criterion for comparing admissible CGPs in terms of informational loss.

The structure of the paper is as follows. Section~\ref{sec2} introduces the basic framework of coarse-grained partitions on finite discrete totally ordered sets. Section~\ref{sec3} formalizes the object-to-category map and the induced probability distributions associated with coarse-grained evaluation. Section~\ref{sec4} defines CU and the KL-based measure of information loss, and proves the zero-information-loss theorem. Section~\ref{sec5} examines the optimization problem of coarse-graining design and discusses its significance for explainable AI and related decision-making contexts. Section~\ref{sec6} concludes.

\section{Coarse-Grained Evaluation}\label{sec2}
\subsection{Coarse-Grained Partitions}\label{sec2.1}
In many institutional settings, normative evaluations are implemented on finite, discrete score sets. For instance, school test scores are recorded as bounded integers rather than real numbers. In this paper, we take such discretized scores as our main motivating case. Accordingly, it suffices to model the fine-grained score domain as an arbitrary finite totally ordered set (e.g., a finite subset of 
$\mathbb{N}$). Throughout the paper, we fix a finite totally ordered set $(U,\leq)$, which we call the underlying scale.
In applications, one may take $U=\{0,1,\dots,100\}$ with the natural order.
We equip $U$ with the discrete $\sigma$-algebra $2^U$.

What must be discussed rigorously, when we speak of coarsening a fine-grained evaluation, is how the resulting coarse-graining should be represented as a mathematical object. Here we formalize it as a coarse-grained partition of $U$, which will later induce a score-to-category map.

\begin{definition}[Coarse-grained partition]\label{def:cgp_discrete}
Given the underlying scale $(U,\le)$, let $\mathscr{I}(U,\le)$ denote the family of all nonempty intervals of $U$.
A \emph{coarse-grained partition} of $(U,\le)$ is a partition object $\pi$ equipped with a finite
index set $I_\pi$ and a family of grains
\[
\mathfrak{G}_\pi := \{G_{\pi,i}\}_{i\in I_\pi}
\]
such that:
(i) each $G_{\pi,i}\in\mathscr{I}(U,\le)$ is a nonempty interval;
(ii) $G_{\pi,i}\cap G_{\pi,j}=\varnothing$ for $i\ne j$;
(iii) $\bigcup_{i\in I_\pi} G_{\pi,i}=U$.
We call each $G_{\pi,i}$ a \emph{grain} (or \emph{category}) of $\pi$.
\end{definition}

\noindent
Here $\pi$ denotes the structure specifying how $U$ is partitioned, while $\mathfrak{G}_\pi$ denotes its family of grains.

\begin{example}
Let $U_1:=\{1,2,3\}$ be a finite totally ordered set with the natural order $\le$.
The nonempty intervals (order-convex subsets) of $(U_1,\le)$ are
\[
\{1\},\ \{2\},\ \{3\},\ \{1,2\},\ \{2,3\},\ \{1,2,3\}.
\]
Equivalently,
\[
\mathscr{I}(U_1,\le)=\bigl\{\{1\},\{2\},\{3\},\{1,2\},\{2,3\},\{1,2,3\}\bigr\}.
\]
Applying Definition~\ref{def:cgp_discrete}, there are exactly four interval
coarse-grained partitions of $U_1$. Writing each partition $\pi_k$ through its
grain family $\mathfrak{G}_{\pi_k}$, we obtain:
\begin{align*}
\mathfrak{G}_{\pi_1} &= \bigl\{\{1\},\{2\},\{3\}\bigr\},\\
\mathfrak{G}_{\pi_2} &= \bigl\{\{1,2\},\{3\}\bigr\},\\
\mathfrak{G}_{\pi_3} &= \bigl\{\{1\},\{2,3\}\bigr\},\\
\mathfrak{G}_{\pi_4} &= \bigl\{\{1,2,3\}\bigr\}.
\end{align*}

In this context, each element of the grain family associated with a partition (e.g., $\{1,2\}$ in $\mathfrak{G}_{\pi_2}$) is a grain (interval) of $U_1$, so this terminology is natural.
\end{example}

%The collection of (interval) coarse-grained partitions of $U$ is denoted by
%$\mathcal{P}_{\textrm{int}}(U)$ in this paper.
%In particular, for the example $U_1=\{1,2,3\}$ above, we have
%\[
%\mathcal{P}_{\textrm{int}}(U_1)=\{\pi_1,\pi_2,\pi_3,\pi_4\},
%\]
%where each partition $\pi_k$ is represented by its grain family $\mathfrak{G}_{\pi_k}$ as listed in the example.
%Accordingly, the collection of grain families induced by interval partitions is
%\[
%\mathscr{G}_{\textrm{int}}(U_1)
%=\{\mathfrak{G}_{\pi_1},\mathfrak{G}_{\pi_2},\mathfrak{G}_{\pi_3},\mathfrak{G}_{\pi_4}\}.
%\]

%\begin{remark}[On generality]\label{rem:generality}
%The discrete ordered framework above is sufficient for the main results of this paper and matches typical applications such as grading scales. A measure-theoretic extension to general measurable spaces can be developed separately (e.g.\ by replacing sums with integrals).
%\end{remark}

\begin{remark}[Ordering grain families]\label{rem:grainorder}
A key structural feature of interval coarse-graining is that it induces an order not only on $U$ but also on the grain family itself.
Define, for $G,H\in\mathfrak{G}_\pi$,
\[
G \preccurlyeq H \quad:\Longleftrightarrow\quad (\forall x\in G)(\forall y\in H)\; x\leq y.
\]
This element-wise relation captures a \emph{global} ordering of categories and is compatible with the intuitive notion that
some coarse-grained categories are uniformly ``lower'' (or ``worse'') than others along the evaluation scale.
Although the present paper focuses on probabilistic pushforwards and information loss under within-grain unification,
this induced grain order provides the missing structure needed to define order-sensitive operators on partitions
(e.g., adjacent merging, monotone coarsening, and order-constrained optimization of coarse-grainings).
\end{remark}

%\begin{remark}[No internal order within grains]\label{rem:no_internal_order}
%In our constructions, each grain $G\in\mathfrak{G}_\pi$ is treated as an \emph{unordered} set: only membership $u\in G$ and the cardinality $|G|$ are used. In particular, the CU construction assigns probabilities uniformly within each grain, so no internal ordering of elements inside a grain is required in this paper. By contrast, in extensions where one selects a representative value for each grain---for instance, taking $\min(G)$ as the representative---the ambient order on $U$ (and hence the induced order within $G$) becomes relevant.
%\end{remark}

If arbitrary set partitions were allowed, then the number of partitions of an $n$-element set would be the Bell number $B_n$,
which grows extremely rapidly.
By contrast, restricting to order-convex (interval) grains makes each partition equivalent to choosing cut points
between adjacent elements; hence there are exactly $2^{\,n-1}$ interval partitions of a finite totally ordered $n$-element set.
This restriction rules out nonlocal groupings while keeping the combinatorial complexity of the partition space
substantially lower than in the unrestricted case.

\begin{proposition}[Counting interval partitions]\label{pro1}
Let $U$ be a finite totally ordered set with $|U|=n$.
Then the number of coarse-grained partitions of $U$ into order-convex grains
(i.e., interval partitions) is
\[
2^{\,n-1}.
\]
\end{proposition}
\begin{proof}[Proof of Proposition \ref{pro1}]
List the elements of $U$ in increasing order as $u_1\prec u_2\prec \cdots \prec u_n$.
An interval partition is uniquely determined by deciding, for each of the $n-1$ gaps
between consecutive elements $(u_i,u_{i+1})$, whether to place a cut (start a new grain)
or not. Each gap offers two choices, independently.
Hence there are $2^{\,n-1}$ interval partitions.
\end{proof}

\section{Object-to-Category Map}\label{sec3}
\subsection{Definition}
CE studies how a fine-grained evaluation on a scale $U$ is mapped into a coarser categorization.
In practice, an evaluation is always performed \emph{of} some collection of objects. 
For example, the evaluation ``Emily is in rank A because she scored 90 on a math test'' can be viewed as a two-step mapping: Emily (as an object) is first mapped to a fine-grained numerical score (90 points), and this score is then mapped to a coarse category (rank A).
Accordingly, we need a map from each object to a category (an object-to-category map); however, this map is the composition of two maps: one from the object to a fine-grained score (an object-to-score map), and another from that fine-grained score to a coarse category (a score-to-category map).

We begin by defining the score-to-category map. Given a coarse-grained partition $\pi$ of $U$ with grain family
$\mathfrak{G}_\pi=\{G_{\pi,i}\}_{i\in I_\pi}$, the coarse categories are conceptually given by the grains
$G_{\pi,i}\in\mathfrak{G}_\pi$.
For technical convenience (measurability and pushforward constructions), we represent these categories
by their labels $I_\pi$ and fix a score-to-category map
\[
q_\pi:U\to I_\pi,\qquad q_\pi(x)=i\ \Longleftrightarrow\ x\in G_{\pi,i}.
\]

%------------------------------------------------
% Index-based coarse assignment and pushforward
%------------------------------------------------

\begin{definition}[Score-to-category map]\label{def:index_assignment}
Let $(U,\le)$ be a finite totally ordered set, equipped with the discrete $\sigma$-algebra $2^U$.
Let $\pi$ be a partition of $U$ with grain family $\mathfrak{G}_\pi=\{G_{\pi,i}\}_{i\in I_\pi}$ (cf.\ Def.~\ref{def:cgp_discrete}).
Define the \emph{index set} $I_\pi$ as the label set appearing in $\mathfrak{G}_\pi=\{G_{\pi,i}\}_{i\in I_\pi}$,
equipped with the discrete $\sigma$-algebra $2^{I_\pi}$.
The \emph{score-to-category map} associated with $\pi$ is the map
$q_\pi:U\to I_\pi$ defined by
\[
q_\pi(x)=i \quad \Longleftrightarrow \quad x\in G_{\pi,i}.
\]
\end{definition}

\begin{example}[A score-to-category map]\label{ex:qtheta}
Let $U=\{1,2,3\}$ with the discrete $\sigma$-algebra $2^U$. Define a coarse-grained partition $\pi$ of $U$
by the grain family
\[
\mathfrak{G}_\pi:=\{G_{\pi,i}\}_{i\in I_\pi}=\bigl\{\{1\},\{2,3\}\bigr\},
\qquad I_\pi:=\{1,2\}.
\]
Define the score-to-category map $q_\pi:U\to I_\pi$ by $q_\pi(u)=i$ iff $u\in G_{\pi,i}$.
Then
\[
q_\pi(1)=1,\qquad q_\pi(2)=2,\qquad q_\pi(3)=2.
\]
\end{example}

%\begin{remark}[Measurability in the finite/discrete setting]\label{rem:measurability_discrete}
%Since we work with the discrete $\sigma$-algebras $2^U$ and $2^{I_\pi}$ on finite sets,
%\emph{every} map between these spaces is measurable. In particular, the score-to-category map
%$q_\pi:U\to I_\pi$ is automatically measurable.
%\end{remark}

\begin{definition}[Object-to-category map]\label{def:obj_to_cat}
Let $O$ be a nonempty set of objects and let $f:O\to U$ be an object-to-score map.
Fix a coarse-grained partition $\pi$ of $U$ with grain family
$\mathfrak{G}_\pi=\{G_{\pi,i}\}_{i\in I_\pi}$ and score-to-category map $q_\pi:U\to I_\pi$.
The \emph{object-to-category map} induced by $f$ and $\pi$ is the composition
\[
q_\pi\circ f:O\to I_\pi,\qquad o\mapsto q_\pi(f(o)).
\]
\end{definition}

\subsection{Illustration by School Test}\label{sec3.3}

Let's consider a specific example. Suppose Olivia, Noah, and James took a mathematics test and scored out of 100 points.
The scores are then converted into letter grades as follows:
0--59 points correspond to \emph{poor}, 60--69 to \emph{fair}, 70--79 to \emph{good}, 80--89 to \emph{very good},
and 90--100 to \emph{excellent}. In this evaluation system, a scale from 0 to 100 is the underlying set $U$,
and the letter grades represent its grains.

Let
\[
U := \{0,1,2,\dots,99,100\}\quad\text{with the natural order } \leq \text{ on } U.
\]
Let $I_\pi:=\{1,2,3,4,5\}$ and define a finite coarse-grained partition $\pi$ of $U$ by
\[
\mathfrak{G}_\pi:=\{G_{\pi,i}\}_{i\in I_\pi},
\]
where the grains are given by
\begin{align*}
G_{\pi,1} &:= \{0,\dots,59\}, &
G_{\pi,2} &:= \{60,\dots,69\},\\
G_{\pi,3} &:= \{70,\dots,79\}, &
G_{\pi,4} &:= \{80,\dots,89\},\\
G_{\pi,5} &:= \{90,\dots,100\}.
\end{align*}
For readability, introduce a label map $\ell:I_\pi\to L$ where
\begin{align*}
L:= & \{\text{poor},\text{fair},\text{good},\text{very good},\text{excellent}\}, \\
& \ell(1)=\text{poor},\ \ell(2)=\text{fair},\ \ell(3)=\text{good},\ \ell(4)=\text{very good},\ \ell(5)=\text{excellent}.
\end{align*}

Then:
\begin{align*}
&O=\{\text{Olivia},\text{Noah},\text{James}\},\\
&f:O \to U;\quad f(\text{Olivia}) = 90,\ f(\text{Noah}) = 71,\ f(\text{James}) = 77,\\
&q_\pi: U \to I_\pi;\quad q_\pi(x)=i \Longleftrightarrow x\in G_{\pi,i},\\
&\ell\circ q_\pi \circ f:O \to L;\\
&\quad (\ell\circ q_\pi \circ f)(\text{Olivia})=\text{excellent},\ 
(\ell\circ q_\pi \circ f)(\text{Noah})=\text{good},\ 
(\ell\circ q_\pi \circ f)(\text{James})=\text{good}.
\end{align*}

\noindent
In this case, comparing the mappings $f:O\to U$ and $\ell\circ q_\pi\circ f:O\to L$ is informative.
While $f$ distinguishes all three individuals by their exact scores, the coarse evaluation collapses distinct scores
(e.g., Noah and James) into the same label ``good''. This loss of within-grain distinction affects the induced
frequencies over labels, as discussed in the next section.

\begin{remark}
The role of the function $q_\pi$ can be understood as follows: if the three students are aware only of the grading system and the results, they perceive their results as multivalued information. For example, if they are told that Olivia received an ``excellent'' grade, they know that her score was one of 90, 91, 92, 93, 94, 95, 96, 97, 98, 99, or 100. In other words, the role of $q_\pi$ is to replace the information that Olivia scored 90 with the information that her score was one of 90 through 100. This means that the original information is lost, which gives practical value to the considerations in Section \ref{sec5}.
\end{remark}

\section{Information Loss}\label{sec4}
\subsection{Probability Measures}\label{sec4.1}
Since the previous section defined the evaluation objects and the maps $f:O\to U$ (from objects to fine-grained scores) and $q_\pi:U\to I_\pi$ (from fine-grained scores to category labels), we now compare the induced distributions on the fine-grained score scale $U$ and on the label set $I_\pi$. Concretely, given a probability distribution on $O$, we obtain the score distribution on $U$ as the pushforward along $f$, and the coarse distribution on $I_\pi$ as the pushforward along $q_\pi\circ f$.

\begin{definition}[Pushforward (coarse-grained) probability]\label{def:pushforward_Ptheta}
Given the underlying scale $(U,\le)$, let $\pi$ be a coarse-grained partition of $(U,\le)$ with grain family
$\mathfrak{G}_\pi=\{G_{\pi,i}\}_{i\in I_\pi}$, and let $P_U$ be a probability distribution on $(U,2^U)$.
With $q_\pi$ as in Definition~\ref{def:index_assignment}, we define the \emph{coarse-grained probability distribution}
$P_\pi$ on $(I_\pi,2^{I_\pi})$ as follows.
First, for each label $i\in I_\pi$, set
\[
P_\pi(\{i\}) := P_U(G_{\pi,i}).
\]
We then extend this to arbitrary subsets $A\subseteq I_\pi$ by additivity, defining
\[
P_\pi(A) := \sum_{i\in A} P_\pi(\{i\})
          = \sum_{i\in A} P_U(G_{\pi,i})
          = P_U\!\left(\bigcup_{i\in A} G_{\pi,i}\right).
\]
Equivalently, for every $A\subseteq I_\pi$,
\[
P_\pi(A) = P_U\bigl(q_\pi^{-1}(A)\bigr),
\]
that is, $P_\pi$ is the pushforward of $P_U$ along $q_\pi$.
\end{definition}

\begin{example}[Pushforward to a pass/fail outcome]\label{ex:pushforward-test}
As a running example, consider now that we are a small classroom teacher and give a math test to ten students. The results appear in Table \ref{tab:student_scores}, Figure \ref{fig1}.

\begin{table}[hbt]
    \centering
    \caption{Student Scores} 
    \label{tab:student_scores}
    \begin{tabular}{|c|c|c|}
        \hline
        \textbf{Student} & \textbf{Score} & \textbf{Pass/Fail}\\
        \hline
        $s_1$ & 5 & pass \\
        $s_2$ & 3 & fail \\
        $s_3$ & 3 & fail \\
        $s_4$ & 3 & fail \\
        $s_5$ & 10 & pass \\
        $s_6$ & 6 & pass \\
        $s_7$ & 2 &fail \\
        $s_8$ & 5 &pass  \\
        $s_9$ & 4 &fail\\
        $s_{10}$ & 0 & fail \\
        \hline
    \end{tabular}
\end{table}

\begin{figure}[hbt]
\centering
\caption{Distribution of Student Scores (Right-Skewed)} 
\includegraphics[scale=0.5]{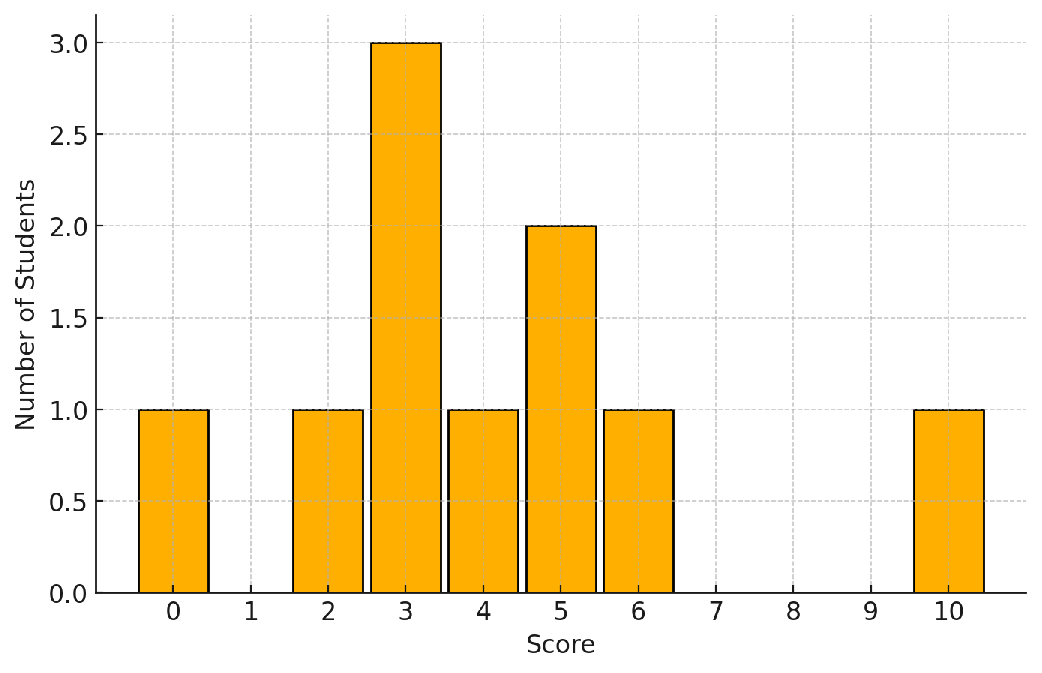}
\label{fig1}
\end{figure}

We can represent the results using CGPs as follows:
Let the underlying score scale be the finite totally ordered set
\[
U:=\{0,1,2,\dots,10\}\quad\text{with the natural order } \le.
\]
Let the set of students be
\[
O:=\{s_1,s_2,\dots,s_{10}\}.
\]
Define the object-to-score map $f:O\to U$ from Table~\ref{tab:student_scores} by
\begin{align*}
& f(s_1)=5,\ f(s_2)=3,\ f(s_3)=3,\ f(s_4)=3,\ f(s_5)=10, \\
& f(s_6)=6,\ f(s_7)=2,\ f(s_8)=5,\ f(s_9)=4,\ f(s_{10})=0.
\end{align*}

\medskip
\noindent
The empirical distribution induced by $f$ is $\hat P\in\Delta(U)$ defined by
\[
\hat P(u):=\frac{\bigl|\{s\in O:\ f(s)=u\}\bigr|}{|O|}\qquad (u\in U).
\]
Hence
\begin{align*}
& \hat P(0)=\frac{1}{10},\ \hat P(1)=0,\ \hat P(2)=\frac{1}{10},\ \hat P(3)=\frac{3}{10},\ \hat P(4)=\frac{1}{10},\ \\
& \hat P(5)=\frac{2}{10},\ \hat P(6)=\frac{1}{10},\ \hat P(7)=0,\ \hat P(8)=0,\ \hat P(9)=0,\ \hat P(10)=\frac{1}{10}.
\end{align*}

\medskip
\noindent
Define a two-grain (interval) partition $\pi$ of $U$ with grains
\[
G_{\pi,1}:=\{0,1,2,3,4\}\quad(\text{fail-grain}),\qquad
G_{\pi,2}:=\{5,6,7,8,9,10\}\quad(\text{pass-grain}),
\]
and let the associated grain family be
\[
\mathfrak{G}_\pi:=\{G_{\pi,1},G_{\pi,2}\}.
\]
Let $I_\pi:=\{1,2\}$ be the index set and let $L:=\{\text{fail},\text{pass}\}$ be the label set.
Define the label map $\ell:I_\pi\to L$ by
\[
\ell(1)=\text{fail},\qquad \ell(2)=\text{pass}.
\]
Define the score-to-category map $q_\pi:U\to I_\pi$ by
\[
q_\pi(u)=
\begin{cases}
1, & u\in G_{\pi,1},\\
2, & u\in G_{\pi,2}.
\end{cases}
\]
Then the labeled coarse evaluation is $\ell\circ q_\pi:U\to L$, and the induced pass/fail labeling on students is
\[
\ell\circ q_\pi\circ f:O\to L.
\]
For the given data,
\begin{align*}
& (\ell\circ q_\pi\circ f)(s_1)=\text{pass},\ 
(\ell\circ q_\pi\circ f)(s_2)=\text{fail},\ 
(\ell\circ q_\pi\circ f)(s_3)=\text{fail},\ \\
& (\ell\circ q_\pi\circ f)(s_4)=\text{fail},\ 
(\ell\circ q_\pi\circ f)(s_5)=\text{pass},
(\ell\circ q_\pi\circ f)(s_6)=\text{pass},\ \\
& (\ell\circ q_\pi\circ f)(s_7)=\text{fail},\ 
(\ell\circ q_\pi\circ f)(s_8)=\text{pass},\ 
(\ell\circ q_\pi\circ f)(s_9)=\text{fail},\ \\
& (\ell\circ q_\pi\circ f)(s_{10})=\text{fail}.
\end{align*}

\medskip
Define the pushforward distribution on labels $\hat P_L$ on $(L,2^L)$ by
\[
\hat P_L(A):=\hat P\bigl((\ell\circ q_\pi)^{-1}(A)\bigr)\qquad (A\subseteq L).
\]
In particular, for a singleton $A=\{\lambda\}$ with $\lambda\in L$, we have
\[
\hat P_L(\{\lambda\})
=\hat P\bigl((\ell\circ q_\pi)^{-1}(\{\lambda\})\bigr)
=\sum_{u\in(\ell\circ q_\pi)^{-1}(\{\lambda\})}\hat P(u).
\]
Thus,
\[
\hat P_L(\text{fail})=\sum_{u=0}^{4}\hat P(u)=\frac{6}{10}=\frac{3}{5},\qquad
\hat P_L(\text{pass})=\sum_{u=5}^{10}\hat P(u)=\frac{4}{10}=\frac{2}{5}.
\]
Equivalently, $\hat P_L(\text{fail})=\hat P(G_{\pi,1})$ and $\hat P_L(\text{pass})=\hat P(G_{\pi,2})$.

\end{example}

\subsection{Kullback--Leibler Divergence}\label{sec4.2}
\subsubsection{Setup}
We quantify the information loss caused by coarse-graining via the Kullback--Leibler (KL) divergence between probability distributions
 due to its well-established role in measuring the divergence between probability distributions in information theory \citep{kullbackleibler1951}. 
\begin{equation}
D_{\text{KL}}(P \parallel Q) = \sum_{i} P(i) \log \frac{P(i)}{Q(i)}
\end{equation}

While alternative measures such as Jensen-Shannon Divergence or Wasserstein Distance may be explored in future studies, $D_\text{KL}$ provides a mathematically straightforward and interpretable baseline for assessing information loss in coarse evaluations.

Using \textit{Example}~\ref{ex:pushforward-test}, we compare the distributions induced by the original 0--10 scoring scheme and by its binary pass/fail coarsening. Intuitively, if only pass/fail outcomes are retained (and the exact scores are discarded), we would like to quantify how much information is lost.

When using $D_{\text{KL}}$ to achieve this objective, determining $P(i)$ is straightforward. The probability $P(i)$ represents the proportion of students who received a specific score $i$ in the original grading system. It is computed as the ratio of the number of students with score $i$ to the total number of students.
 It involves assigning the proportion of elements in $O$ that map to each element in $U$. For instance, $P(0)$ is $\frac{1}{10}$, and $P(1)$ is 0 as to Table \ref{tab:student_scores}.

\subsubsection{Categorical Unification}\label{sec4.3}
The challenge lies in determining $Q(i)$. A straightforward but probably unrealistic assumption would be to assign equal probability to all scores (Table \ref{tab:unification_based_qi}). Under the assumption that $Q(i)$ follows a uniform distribution ($Q(i) = \frac{1}{11}$ for all $i$), the $D_{\text{KL}}$ is calculated to be approximately 0.564. This value quantifies the extent to which the uniform assumption deviates from the actual probability distribution $P(i)$.

\begin{table}[hbt]
    \centering
    \caption{Unification of $Q(i)$} 
    \begin{tabular}{|c|c|c|}
        \hline
        \textbf{Score} & \textbf{Original Probability $P(i)$} & \textbf{Coarse-grained $Q(i)$} \\
        \hline
        0 & $\frac{1}{10}$ & $\frac{1}{11}$ \\
        1 & $0$ & $\frac{1}{11}$ \\
        2 & $\frac{1}{10}$ & $\frac{1}{11}$ \\
        3 & $\frac{3}{10}$ & $\frac{1}{11}$ \\
        4 & $\frac{1}{10}$ & $\frac{1}{11}$ \\
        5 & $\frac{2}{10}$ & $\frac{1}{11}$ \\
        6 & $\frac{1}{10}$ & $\frac{1}{11}$ \\
        7 & $0$ & $\frac{1}{11}$ \\
        8 & $0$ & $\frac{1}{11}$ \\
        9 & $0$ & $\frac{1}{11}$ \\
        10 & $\frac{1}{10}$ & $\frac{1}{11}$ \\
        \hline
    \end{tabular}
    \label{tab:unification_based_qi}
\end{table} 

However, this assumption is generally unrealistic: even among passing students, scores within the
same category (grain) need not be evenly distributed. At the same time, once we coarse-grain the
score scale, we want the coarsened distribution to preserve the total probability mass assigned to
each grain, and we would like to define a canonical representative distribution that introduces no
additional structure beyond this coarse information. To this end, we define \textit{Categorical Unification} (CU) as follows.

\begin{definition}[Categorical unification and $D_{\mathrm{KL\text{-}CU}}$]\label{def:CU_discrete}
Given the underlying scale $(U,\le)$, let $P_U$ be a probability distribution on $(U,2^U)$.

\smallskip
\noindent
(i) The \emph{coarse-grained distribution} induced by $\pi$ is the probability distribution
$P_\pi$ on $I_\pi$ defined by
\[
P_\pi(\{i\}) := P_U(G_{\pi,i})=\sum_{u\in G_{\pi,i}} P_U(\{u\}),\qquad i\in I_\pi.
\]
Equivalently, in the discrete setting we write $P_U(u):=P_U(\{u\})$ and $P_\pi(i):=P_\pi(\{i\})$.

\smallskip
\noindent
(ii) The \emph{CU-unification} of $P_U$ (via $\pi$) is the probability distribution
$Q^{\mathrm{CU}}$ on $U$ defined by
\[
Q^{\mathrm{CU}}(u):=\frac{P_\pi(\{i\})}{|G_{\pi,i}|}\qquad\text{for }u\in G_{\pi,i}.
\]
Since $\pi$ is a partition, each $u\in U$ belongs to a unique grain $G_{\pi,i}$, so $Q^{\mathrm{CU}}$ is well-defined.
Note that $Q^{\mathrm{CU}}$ is indeed a probability distribution on $U$: it redistributes each coarse mass
$P_\pi(\{i\})$ uniformly over the points $u\in G_{\pi,i}$. In particular,
\[
\sum_{u\in U} Q^{\mathrm{CU}}(u)
=\sum_{i\in I_\pi}\sum_{u\in G_{\pi,i}} \frac{P_\pi(\{i\})}{|G_{\pi,i}|}
=\sum_{i\in I_\pi} P_\pi(\{i\})
=1.
\]

\smallskip
\noindent
(iii) The \emph{KL divergence by categorical unification} is
\[
D_{\mathrm{KL\text{-}CU}}(P_U):=D_{\mathrm{KL}}\!\left(P_U\,\middle\|\,Q^{\mathrm{CU}}\right),
\]
where
\[
D_{\mathrm{KL}}(P\|Q)=\sum_{u\in U} P(u)\log\!\frac{P(u)}{Q(u)},
\]
with the convention that any term with $P(u)=0$ is taken to be $0$.
\end{definition}

The point of Definition~\ref{def:CU_discrete} is not to compare the original distribution directly with the coarse representation itself. Rather, $D_{\mathrm{KL\text{-}CU}}$ measures how far the original distribution deviates from a canonical fine-scale reconstruction induced by the coarse representation under minimal assumptions.

The empirical rationale for this method is as follows: When students are evaluated using a coarse-grained grading system where only categorical results (e.g., pass or fail) are recorded, their exact scores become irretrievable. If a student is classified as ``pass'', the only information available is that their score fell within the pass range (e.g., 5 to 10). Without further data, we cannot determine if they barely passed or aced the test. Given this limitation, assigning equal probability to all scores within the pass category is a reasonable and minimally biased assumption. Since no particular score within the pass category has more supporting evidence than another, a uniform assumption ensures neutrality and avoids introducing arbitrary weighting.

Additionally, the principle of maximum entropy provides a theoretical justification for CU. It states that among all probability distributions satisfying a given set of constraints, the one that maximizes entropy (i.e., uncertainty) is the least biased and least informative beyond the imposed constraints \citep{jaynes1957}. In the case of pass/fail classification, however, we are not only constrained by the requirement that probabilities sum to one but also by the fact that we know the number of students in each category. Specifically, there are six students in the fail category and four in the pass category, meaning that the total probability for each category is $\frac{6}{10}$ and $\frac{4}{10}$, respectively. According to the principle of maximum entropy, the probability distribution within each category should be uniform to maximize entropy. That is, within the fail category, the assumption that maximizes entropy (and therefore least biased) is a uniform distribution among the five elements (i.e., 0 to 4), and similarly, a uniform distribution should be assumed within the pass category (i.e., 5 to 10). Therefore, CU assigns uniform probability within each category, which corresponds to the entropy-maximizing distribution given the category-level constraints (Table \ref{tab:density_based_qi}). Since an entropy-maximizing distribution is the one that assumes the least additional information, CU can be considered the most unbiased method for probability assignment within coarse-grained categories.

\begin{proposition}[Maximum-entropy property of CU]\label{prop:CU_maxent}
Given the underlying scale $(U, \leq)$, fix $P\in\Delta(U)$ and write
\[
p_i := P(G_{\pi,i})=\sum_{x\in G_{\pi,i}} P(x)\qquad (i\in I_\pi).
\]
Define the CU of $P$ (relative to $\pi$) by
\[
P^{\mathrm{CU}}(x):=\frac{p_i}{|G_{\pi,i}|}\quad\text{for }x\in G_{\pi,i}.
\]
Consider the constraint set
\[
\mathcal{C}(p,\pi):=\Bigl\{Q\in\Delta(U)\ \bigm|\ Q(G_{\pi,i})=p_i\text{ for all }i\in I_\pi\Bigr\}.
\]
Then $P^{\mathrm{CU}}$ maximizes Shannon entropy $H(Q):=-\sum_{x\in U}Q(x)\log Q(x)$ over $\mathcal{C}(p,\pi)$.
Moreover, if $p_i>0$ for all $i$, the maximizer is unique.
\end{proposition}

The constraint set $\mathcal{C}(p,\pi)$ fixes only the coarse information, namely the total mass
$p_i$ assigned to each grain $G_{\pi,i}$, while leaving the distribution within each grain free.
For example, if a grain is $G_{\pi,i}=\{8,9,10\}$ and $p_i=0.3$, then any assignment of nonnegative
probabilities $(Q(8),Q(9),Q(10))$ with $Q(8)+Q(9)+Q(10)=0.3$ is admissible, such as $(0.1,0.1,0.1)$ or
$(0.29,0.01,0)$.
Proposition~\ref{prop:CU_maxent} shows that, among all fine-grained distributions consistent with these
coarse masses, the CU distribution $P^{\mathrm{CU}}$ is the least informative one in the sense of
maximum Shannon entropy (in the above example, this corresponds to the uniform assignment $(0.1,0.1,0.1)$
on $G_{\pi,i}=\{8,9,10\}$): it preserves the grain-level totals $p_i$ and otherwise spreads mass uniformly
within each grain. In particular, CU can be viewed as the canonical ``uninformative'' refinement of the
coarse-grained distribution back to the original scale $U$.

\begin{proof}[Proof of Proposition \ref{prop:CU_maxent}]
Fix any $Q\in\mathcal{C}(p,\pi)$. For each $i$ with $p_i>0$, define the conditional distribution on the grain
\[
Q_i(x):=\frac{Q(x)}{p_i}\quad (x\in G_{\pi,i}),
\]
so that $\sum_{x\in G_{\pi,i}}Q_i(x)=1$.
A direct calculation yields the entropy decomposition
\[
H(Q)=H(p)+\sum_{i\in I_\pi} p_i\, H(Q_i).
\]
This identity shows that the total entropy of $Q$ consists of the entropy $H(p)$ at the coarse-grained level (which grain is selected) plus the average within-grain entropy $\sum_{i\in I_\pi} p_i H(Q_i)$, weighted by the grain masses $p_i$. $H(p):=-\sum_i p_i\log p_i$ depends only on $p$ (hence is constant over $\mathcal{C}(p,\pi)$)
Therefore, maximizing $H(Q)$ over $\mathcal{C}(p,\pi)$ reduces to maximizing each $H(Q_i)$ over the simplex
$\Delta(G_{\pi,i})$.
For a fixed finite set $G_{\pi,i}$, Shannon entropy is maximized uniquely by the uniform distribution on $G_{\pi,i}$,
i.e., $Q_i(x)=1/|G_{\pi,i}|$ for all $x\in G_{\pi,i}$.
Thus, for all $x\in G_{\pi,i}$,
\[
Q(x)=p_i Q_i(x)=\frac{p_i}{|G_{\pi,i}|}=P^{\mathrm{CU}}(x),
\]
so $P^{\mathrm{CU}}$ is the (unique, if all $p_i>0$) maximizer.
\end{proof}

\begin{table}[hbt]
    \centering
    \caption{Categorical Unification of $Q(i): T=5$}
    \begin{tabular}{|c|c|c|c|}
        \hline
        \textbf{Score} & \textbf{Original Probability $P(i)$} & \textbf{Pass/Fail} & \textbf{Coarse-grained $Q(i)$} \\
        \hline
        0 & $\frac{1}{10}$ & Fail & $\frac{6}{10} \cdot \frac{1}{5} = \frac{3}{25}$ \\
        1 & $0$ &  Fail & $\frac{3}{25}$ \\
        2 & $\frac{1}{10}$ & Fail & $\frac{3}{25}$ \\
        3 & $\frac{3}{10}$ &  Fail & $\frac{3}{25}$ \\
        4 & $\frac{1}{10}$ &  Fail & $\frac{3}{25}$ \\
        5 & $\frac{2}{10}$ &  Pass & $\frac{4}{10} \cdot \frac{1}{6} = \frac{1}{15}$ \\
        6 & $\frac{1}{10}$ &  Pass & $\frac{1}{15}$ \\
        7 & $0$ &  Pass & $\frac{1}{15}$ \\
        8 & $0$ &  Pass & $\frac{1}{15}$ \\
        9 & $0$ &  Pass & $\frac{1}{15}$ \\
        10 & $\frac{1}{10}$ &  Pass & $\frac{1}{15}$ \\
        \hline
    \end{tabular}
    \label{tab:density_based_qi}
\end{table}

\begin{theorem}[Zero information loss under CU (finite/discrete)]\label{thm:zero_KL_CU_discrete}
Given the underlying scale $(U,\leq)$, let $P_U$ be a probability distribution on $U$, and define the induced coarse-grained distribution
\[
P_\pi(i):=\sum_{u\in G_{\pi,i}} P_U(u),\qquad i\in I_\pi.
\]
Define the $Q^{\mathrm{CU}}$ on $U$ by
\[
Q^{\mathrm{CU}}(u):=\frac{P_\pi(i)}{|G_{\pi,i}|}\qquad\text{for }u\in G_{\pi,i}.
\]
Then the discrete KL divergence
\[
D_{\mathrm{KL}}\!\left(P_U\,\middle\|\,Q^{\mathrm{CU}}\right)
:=\sum_{u\in U} P_U(u)\log\frac{P_U(u)}{Q^{\mathrm{CU}}(u)}
\]
is well-defined and satisfies
\[
D_{\mathrm{KL}}\!\left(P_U\,\middle\|\,Q^{\mathrm{CU}}\right)=0
\quad\Longleftrightarrow\quad
\forall i\in I_\pi,\ \forall u\in G_{\pi,i}:\ 
P_U(u)=\frac{P_\pi(i)}{|G_{\pi,i}|}.
\]
Equivalently, for every $i\in I_\pi$ with $P_\pi(i)>0$ and every $A\subseteq G_{\pi,i}$,
\[
P_U(A\mid G_{\pi,i})=\frac{|A|}{|G_{\pi,i}|}.
\]
\end{theorem}

\begin{proof}[Proof of Theorem \ref{thm:zero_KL_CU_discrete}]
First note that $D_{\mathrm{KL}}(P_U\|Q^{\mathrm{CU}})$ is well-defined.
Indeed, if $Q^{\mathrm{CU}}(u)=0$ for some $u\in G_{\pi,i}$, then $P_\pi(i)=0$.
But $P_\pi(i)=\sum_{v\in G_{\pi,i}}P_U(v)=0$ implies $P_U(v)=0$ for all $v\in G_{\pi,i}$, hence $P_U(u)=0$.
Therefore $Q^{\mathrm{CU}}(u)=0$ never occurs at a point with $P_U(u)>0$.

Now recall the standard fact for the discrete KL divergence: for distributions $P$ and $Q$ on a finite set,
\[
D_{\mathrm{KL}}(P\|Q)\ge 0,
\quad\text{and}\quad
D_{\mathrm{KL}}(P\|Q)=0 \Longleftrightarrow P=Q.
\]
Applying this with $P=P_U$ and $Q=Q^{\mathrm{CU}}$, we obtain
\[
D_{\mathrm{KL}}\!\left(P_U\,\middle\|\,Q^{\mathrm{CU}}\right)=0
\Longleftrightarrow
\forall u\in U:\ P_U(u)=Q^{\mathrm{CU}}(u).
\]
By the definition of $Q^{\mathrm{CU}}$, this is equivalent to
\[
\forall i\in I_\pi,\ \forall u\in G_{\pi,i}:\
P_U(u)=\frac{P_\pi(i)}{|G_{\pi,i}|},
\]
i.e., $P_U$ is constant on each grain and equals the grain mass divided equally among its elements.
Finally, for any $i$ with $P_\pi(i)>0$, the conditional distribution
$P_U(u\mid G_{\pi,i}):=P_U(u)/P_\pi(i)$ (for $u\in G_{\pi,i}$) satisfies
\[
P_U(u\mid G_{\pi,i})=\frac{1}{|G_{\pi,i}|},
\]
which is precisely $\mathrm{Unif}(G_{\pi,i})$.
\end{proof}

This theorem characterizes the situation in which elements that are mapped to the same grain (that is, the same category) after coarsening are already assigned uniform probability mass within that grain under the pre-coarsening distribution $P_U$.

\subsection{Optimizing Information Loss}\label{sec4.3}
\subsubsection{Minimizing the KL Divergence}\label{sec4.3.1}
When moving from simple examples such as pass/fail classification to the general design of coarse-grained evaluations, an additional question arises: how should we choose among multiple possible coarse-grainings? Once admissible coarse-grainings have been restricted to CGPs satisfying the structural requirements of CE, the remaining question is how to compare multiple admissible partitions. The role of $D_\mathrm{{KL\text{-}CU}}$ in this section is to provide one such comparative criterion.

As the zero-loss theorem for $D_{\mathrm{KL\text{-}CU}}$ shows, making information loss vanish in the sense of this measure requires that the original fine-grained distribution already be uniform within each grain. In practice, however, the criteria for coarsening an evaluation are rarely chosen so as to align with the original distribution in this way. For example, in grading a mathematics test, a teacher would not normally create a category simply because the numbers of students scoring 61, 62, and 63 happen to be equal. The zero-loss theorem therefore should not be understood as providing a practical guideline for optimizing categorization. Rather, it shows that practical categorization is usually guided by considerations other than the complete elimination of information loss in the sense of $D_{\mathrm{KL\text{-}CU}}$.

\begin{figure}[hbt]
\centering
\caption{Graphing of Table \ref{tab:cu_qi}} 
\includegraphics[scale=0.48]{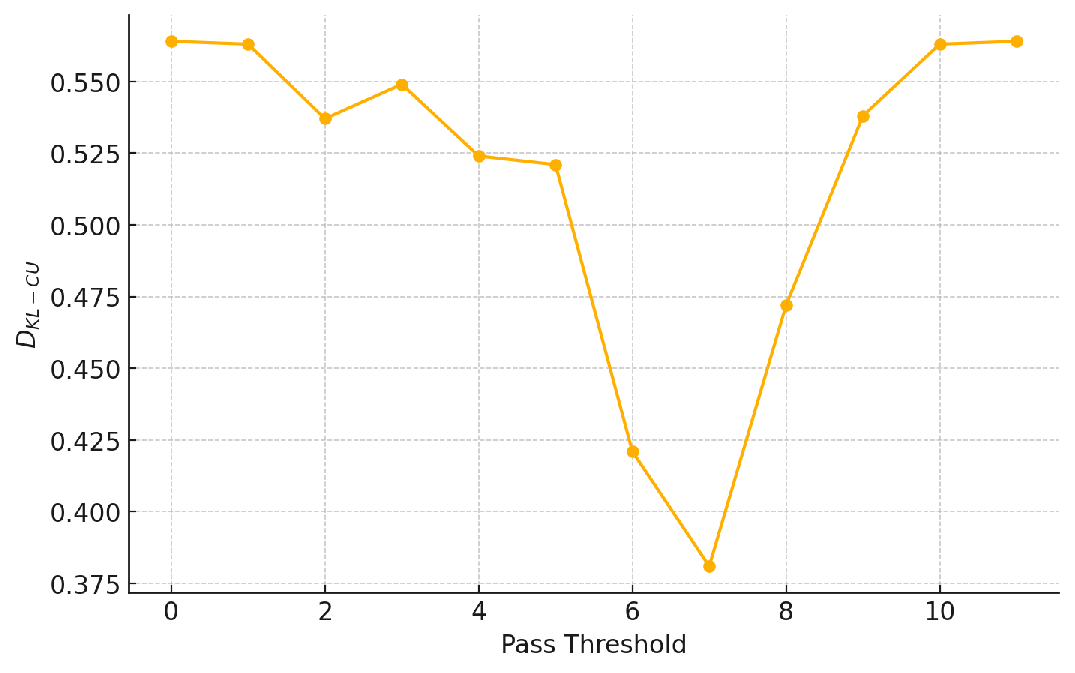}
\label{fig:cu_qi}
\end{figure}

One example in which zero information loss cannot be expected is when the number of categories is fixed in advance—for instance, when outcomes must be classified only as pass or fail. One possible idea, then, is to give up on achieving zero information loss and instead set the threshold so as to minimize information loss. Suppose we need to divide those ten students (Table \ref{tab:student_scores}) into pass and fail groups, but we can determine the passing threshold $(T)$. What score should be used as $T$? In this setting, when the number of grains is fixed, minimizing information loss provides a natural yet straightforward way to formulate the objective function. The lowest $D_{\text{KL-CU}}$ is achieved when the passing threshold is set at $T = 7$, resulting in a value of 0.381 (Table \ref{tab:cu_qi}, Figure \ref{fig:cu_qi}).

\begin{table}[hbt]
    \centering
    \caption{$D_{\text{KL-CU}}$ of Table \ref{tab:student_scores}} 
    \begin{tabular}{|c|c|c|}
        \hline
        \textbf{Pass Threshold} & \textbf{Pass Students} & \textbf{$D_{\text{KL-CU}}$} \\
        \hline
        0 & all students & 0.564 \\
        1 & $s_1,s_2,s_3,s_4,s_5,s_6,s_7,s_8,s_9$ & 0.563 \\
        2 & $s_1,s_2,s_3,s_4,s_5,s_6,s_7,s_8,s_9$ & 0.537 \\
        3 & $s_1,s_2,s_3,s_4,s_5,s_6,s_8,s_9$ & 0.549 \\
        4 & $s_1,s_5,s_6,s_8,s_9$ & 0.524 \\
        5 & $s_1,s_5,s_6,s_8$ & 0.521 \\
        6 & $s_5,s_6$ & 0.421 \\
        7 & $s_5$ & 0.381 \\
        8 & $s_5$ & 0.472 \\
        9 & $s_5$ & 0.538  \\
        10 & $s_5$ & 0.563 \\
        11 & no student & 0.564 \\
        \hline
    \end{tabular}
    \label{tab:cu_qi}
\end{table}

However, reducing the coarse-graining design problem to mere information-loss minimization is not always adequate. In many settings (e.g., pass/fail decisions), the relevant objective is not to reproduce the original score distribution, but rather to minimize decision costs (such as false positives and false negatives) or to optimize expected utility. Consider again the arithmetic test example above. If the pass threshold $T$ is chosen so as to minimize $D_{\mathrm{KL\text{-}CU}}$, the optimum may be a threshold such as $T=7$, under which only the student $s_5$ passes. This is reasonable from the viewpoint of distributional fidelity, because it preserves, within the coarse pass/fail representation, the fact that $s_5$ stands out as an exceptionally strong performer in this class of ten students.

Yet such an information-loss-minimization principle is justified only when the teacher's primary aim is to preserve information about the original score distribution. In educational practice, pass/fail evaluations are often tied instead to an operational criterion, such as whether a student can keep up with the next course module. For example, if scoring at least 6 points is sufficient to follow the next class, the teacher may reasonably wish to set the pass threshold at $T=6$. In that case, selecting $T=7$ by minimizing $D_{\mathrm{KL\text{-}CU}}$ may still be meaningful from the viewpoint of distributional fidelity, but it fails to reflect the actual decision objective and therefore becomes an inappropriate design principle for the coarse-graining.

\subsubsection{An Optimization View of Coarse-Grained Design}
Therefore, KL-based information-loss minimization should be understood as a principled baseline objective, one that may need to be supplemented or replaced when decision-theoretic or normative constraints are dominant. Formally, given a finite totally ordered set $(U,\le)$ and a distribution $P$ on $U$, we consider the optimization problem
\begin{equation}\label{eq:opt_partition_general}
\min_{\pi\in \mathcal{P}_{\mathrm{int}}(U,\le)}
\;\; D_{\mathrm{KL}}\!\left(P \,\middle\|\, Q^{\mathrm{CU}}_{\pi}\right)
\;+\;\lambda\,\Omega(\pi),
\end{equation}
where $\mathcal{P}_{\mathrm{int}}(U,\le)$ denotes the collection of interval coarse-grained partitions,
$Q^{\mathrm{CU}}_{\pi}$ is the CU-unification associated with $\pi$,
$\Omega(\pi)$ is a complexity penalty (e.g., the number of grains $m(\pi)$ or a description-length proxy),
and $\lambda\ge 0$ controls the trade-off between informational fidelity and simplicity.
In this formulation, coarse-graining is not merely an evaluative act, but also the choice of a representation guided by optimization, and one that can be adapted to domain-specific constraints and interpretability demands.

The case $\lambda=0$ corresponds to the limiting situation in which minimizing $D_{\mathrm{KL\text{-}CU}}$ is itself the sole objective. In that case, the finest partition is trivially optimal, since no penalty is imposed on excessive refinement. The present framework is intended primarily for the nontrivial case $\lambda>0$, where the design of coarse-graining requires a trade-off between informational fidelity and coarsening cost. From this perspective, the role of $\lambda$ is not merely technical, but expresses the practical fact that coarse-grained evaluation is introduced precisely because interpretability, simplicity, and other costs of fine-grained representation cannot be ignored.

We leave a systematic study of \eqref{eq:opt_partition_general}, including computational aspects and constraint variants, to future work.

\section{Discussion}\label{sec5}
We show that the core ideas of CE can be given a precise mathematical formulation. Coarse evaluations in everyday life---for example, the statement that ``Olivia received an excellent grade on her mathematics test''---exhibit a clear mathematical structure within the framework of coarse-grained partitions. The fact that an evaluation is coarse---for instance, that Olivia's exact score difference from other students who also received an ``excellent'' rating is intentionally ignored---does not mean that the evaluation is ambiguous. In this sense, CGPs make it possible to distinguish clearly between coarseness and ambiguity in assessment. Therefore, CE's attempt to distinguish coarse evaluations from arbitrary or haphazard evaluations can be justified mathematically.

CE has proposed, as a criterion for comparing fine-grained and coarse-grained evaluations, that a coarse-grained evaluation should not reverse the value ordering embodied in the fine-grained one. However, as the previous literature itself acknowledges, multiple coarse-grained evaluations can satisfy this condition for the same fine-grained evaluation. The problem, therefore, is that this criterion alone cannot sufficiently determine how the coarsening should be carried out. In response, we have proposed a new criterion based on comparing the probability distributions associated with the fine-grained and coarse-grained evaluations, and on minimizing the sum of information loss and coarsening cost. When the fine-grained evaluation is given as a finite discrete totally ordered set, a coarse-grained evaluation constructed within the framework of CGPs allows the corresponding probability distributions and information loss to be defined and computed rigorously. Moreover, the central result of this paper—the zero-loss theorem for $D_{\mathrm{KL\text{-}CU}}$—shows that coarsening yields zero information loss in the sense of this measure only when the original fine-grained distribution is already uniform within each grain. Hence, coarse-grained evaluations with zero information loss, in the sense of $D_{\mathrm{KL\text{-}CU}}$, occur only in highly exceptional cases in actual evaluative practice.

These observations are also relevant to explainable AI. In many AI systems, the model first produces a fine-grained internal evaluation of its target. This is true, for example, in AI-assisted driving systems that compute detailed risk scores, in medical AI that estimates diagnostic likelihoods, and in hiring or interview AI that generates fine-grained personnel assessments. From this perspective, one important problem in XAI is how such fine-grained internal evaluations should be transformed into coarser outputs that are intelligible and usable for human agents.
A simple example is AI-assisted driving. Suppose that the system internally represents driving risk on a fine-grained scale such as $U=\{0,1,\dots,100\}$, while the human-facing interface must display only a small number of warnings, such as \emph{safe}, \emph{caution}, and \emph{danger}. In this case, a CGP represents the transformation from the fine-grained risk scale to these coarse warning categories. The corresponding quantity $D_{\mathrm{KL\text{-}CU}}$ then measures how far the original fine-grained risk distribution deviates from the canonical fine-scale reconstruction induced by that coarse warning system under minimal assumptions.
This makes the design problem more precise. The issue is not only whether the warning system remains intelligible to the driver, but also how much fine-grained risk information is lost when the internal evaluation is compressed into a small number of human-interpretable categories. In such cases, the framework proposed in this paper becomes directly relevant: a principled warning system should balance informational fidelity against the coarsening cost associated with human cognitive burden, instruction complexity, and response usability.

\section{Conclusion}\label{sec6}

This paper has provided a rigorous mathematical framework for analyzing coarse-grained evaluation in the spirit of Coarse Ethics (CE). By introducing coarse-grained partitions (CGPs) on finite discrete totally ordered sets, we showed that coarse evaluations need not be treated as vague or arbitrary, but can instead be represented and analyzed in a precise mathematical form.

Our main contribution is threefold. First, we formalized coarse-grained evaluation as a mapping from a fine-grained score scale to a coarser categorical scale. Second, we introduced categorical unification (CU) and thereby defined a KL-based measure, $D_{\mathrm{KL\text{-}CU}}$, of how far the original fine-grained distribution departs from a canonical fine-scale reconstruction induced by the coarse representation under minimal assumptions. Third, we proved that this divergence vanishes only under a highly restrictive condition: within each grain, the original fine-grained distribution must already be uniform. Hence, zero information loss in the sense of $D_{\mathrm{KL\text{-}CU}}$ is not a realistic benchmark for ordinary evaluative practice, but rather a highly exceptional limiting case.

These results also lead naturally to the problem of optimization. Since multiple coarse-grained evaluations may be compatible with the same fine-grained evaluation, the issue is not merely whether coarse-graining is possible, but how it should be designed. Our analysis suggests that minimizing KL-based information loss provides a principled baseline objective, although in real settings this objective may need to be supplemented or replaced by operational, normative, or decision-theoretic considerations. In this sense, the framework developed here should be understood not as a universal rule for categorization, but as a formal basis for comparing alternative design objectives.

Finally, the framework is also relevant to explainable AI. In AI-assisted settings, a system may hold a fine-grained internal evaluation while needing to present a coarser representation that is intelligible to human users. This can be understood as a transformation from a fine-grained evaluation to a coarse-grained one, and our framework provides a natural way to study the trade-off between preserving critical information and reducing coarsening costs such as cognitive burden and explanatory complexity. Future work should extend the present analysis beyond finite discrete ordered settings and investigate richer optimization problems under domain-specific constraints.

\appendix
\section*{Appendix A. Python Code for Calculating $D_{\text{KL-CU}}$}

\begin{lstlisting}[language=Python, caption = Python Code for $D_\text{KL-CU}$ of Table \ref{tab:cu_qi}, label = code1]
import numpy as np
import matplotlib.pyplot as plt
from collections import Counter

def compute_probability_distribution(scores):
    """Compute P(i)=freq(i)/N for scores 0..max(scores)."""
    total = len(scores)
    max_score = max(scores)
    # Include missing scores with 0 frequency
    freq = Counter(scores)
    p = {i: freq.get(i, 0) / total for i in range(max_score + 1)}
    return p

def coarse_grain_distribution(p, threshold):
    """Group p by T and distribute probability uniformly."""
    scores = sorted(p.keys())
    fail_group = [s for s in scores if s < threshold]
    pass_group = [s for s in scores if s >= threshold]
    
    q = {}
    if fail_group:
        total_fail = sum(p[s] for s in fail_group)
        for s in fail_group:
            q[s] = total_fail / len(fail_group)
    if pass_group:
        total_pass = sum(p[s] for s in pass_group)
        for s in pass_group:
            q[s] = total_pass / len(pass_group)
    return q

def kl_divergence(p, q):
    """Compute D_KL(P||Q)=sum P(i)*log(P(i)/Q(i))."""
    dkl = 0
    for s in p:
        if p[s] > 0 and q.get(s, 0) > 0:
            dkl += p[s] * np.log(p[s] / q[s])
    return dkl

def analyze_thresholds(scores):
    """Compute P and for T compute Q and D_KL.
    Returns list of (T, D_KL)."""
    p = compute_probability_distribution(scores)
    max_score = max(scores)
    results = []
    # Iterate T from 0 to max_score+1
    for T in range(max_score + 2):
        q = coarse_grain_distribution(p, T)
        dkl = kl_divergence(p, q)
        results.append((T, dkl))
    return results

# Test data
test_scores = [5, 3, 3, 3, 10, 6, 2, 5, 4, 0]

# Compute and print KL divergence for each threshold T
results = analyze_thresholds(test_scores)
for T, dkl in results:
    print(f"T = {T:2d}  -->  D_KL(P||Q) = {dkl:.5f}")

# Find optimal threshold with minimal KL divergence
best_threshold, min_dkl = min(results, key=lambda x: x[1])
print(
    f"\nOptimal threshold T = {best_threshold}, "
    f"minimum KL divergence is approx. {min_dkl:.5f}"
)

# Plot the results (optional)
Ts, dkl_vals = zip(*results)
plt.figure(figsize=(8, 5))
plt.plot(Ts, dkl_vals, marker='o', linestyle='-')
plt.xlabel('Pass/Fail Threshold (T)')
plt.ylabel('KL Divergence D_KL(P||Q)')
plt.title('KL Divergence for Each Threshold')
plt.grid(True)
plt.show()
\end{lstlisting}

\section*{Declarations}

\textbf{Funding} \\
This research received no specific grant from any funding agency in the public, commercial, or not-for-profit sectors.

\noindent
\textbf{Conflicts of Interest} \\
The author declares no conflict of interest.

\noindent
\textbf{Ethical Approval} \\
This article does not contain any studies with human participants or animals performed by any of the authors.

\noindent
\textbf{Data Availability} \\
No empirical data was used in this study.

\noindent
\textbf{Code Availability} \\
The Python code used to compute the KL Divergence is available at \url{https://github.com/Takashi-Izumo/CGPs-implementation}.

\bibliography{references}

\end{document}